\documentclass{article}
\usepackage{listings}
\usepackage{adjustbox}
\usepackage{newtxtext,newtxmath}
\usepackage{multirow}
\usepackage{booktabs}
\usepackage{colortbl}
\usepackage{float}           
\usepackage{subfig}                 
\usepackage{authblk}
\usepackage{pdfpages}
\usepackage{hyperref}
\date{}

\title{Customizing General-Purpose Foundation Models for Medical Report Generation}

\author[1,2]{Bang Yang}
\author[2]{Asif Raza}
\author[2]{Yuexian Zou}
\author[1*]{Tong Zhang}
\affil[1]{Peng Cheng Laboratory, Shenzhen 518055, China}
\affil[2]{ADSPLAB, School of Electronic and Computer Engineering, Peking University, Shenzhen 518055, China}
\affil[*]{Correspondence to: zhangt02@pcl.ac.cn}

\begin{document}
\maketitle
\begin{abstract}
    Medical caption prediction which can be regarded as a task of medical report generation (MRG), requires the automatic generation of coherent and accurate captions for the given medical images. 
     However, the scarcity of labelled medical image-report pairs presents great challenges in the development of deep and large-scale neural networks capable of harnessing the potential artificial general intelligence power like large language models (LLMs).
    In this work, we propose customizing off-the-shelf general-purpose large-scale pre-trained models, i.e., foundation models (FMs), in computer vision and natural language processing with a specific focus on medical report generation.
    Specifically, following BLIP-2, a state-of-the-art vision-language pre-training approach, we introduce our encoder-decoder-based MRG model. This model utilizes a lightweight query Transformer to connect two FMs: the giant vision Transformer EVA-ViT-g and a bilingual LLM trained to align with human intentions (referred to as ChatGLM-6B). Furthermore, we conduct ablative experiments on the trainable components of the model to identify the crucial factors for effective transfer learning. Our findings demonstrate that unfreezing EVA-ViT-g to learn medical image representations, followed by parameter-efficient training of ChatGLM-6B to capture the writing styles of medical reports, is essential for achieving optimal results.
    Our best attempt (PCLmed Team) achieved the 4$^{th}$ and the 2$^{nd}$, respectively, out of 13 participating teams, based on the BERTScore and ROUGE-1 metrics, in the ImageCLEFmedical Caption 2023 Caption Prediction Task competition.
\end{abstract}
\section{Introduction}

Medical report generation (MRG) requires the automatic generation of accurate and fluent reports that describe the impressions and findings of medical images, making it promising to address the well-documented ``physician burnout'' phenomenon~\cite{tawfik2018physician,west2018physician}. 
As one of generative medical tasks~\cite{chaves2022automatic,liu2022retrieve,messina2022survey,liu2022graph} and a vision-language (VL) task, MRG needs high-quality cross-modal annotations, i.e., medical image-report pairs. Although a few open-source datasets have been introduced to foster research in this direction~\cite{demner2016preparing,johnson2019mimic,ImageCLEFmedicalCaptionOverview2023},
their data scale, i.e., up to hundreds of thousand of pairs, pose great challenges for developing deep and giant neural networks.

Recently, an increasing number of large-scaled pre-trained models, i.e., foundation models (FMs), are emerging in computer vision (CV) \cite{kirillov2023segment,fang2023eva,wang2023internimage}, natural language processing (NLP)~\cite{ouyang2022training,zeng2023glm,touvron2023llama}, and their intersection~\cite{li2023blip,openai2023gpt4,wu2023visual}. These foundation models are built for general purposes and can handle a wide range of tasks by prompt engineering~\cite{brown2020language,Wei2022CoT,kojima2022CoT} or parameter-efficient transfer learning~\cite{houlsby2019parameter,li2021prefix,hulora,liu2022p}. More importantly, due to model and data scaling, FMs may acquire \emph{emergent abilities} to solve tasks that are difficult for small models~\cite{wei2022emergent}. Given the shortage of labeled medical image-report pairs and the impressive powers of FMs, an intermediate problem rises up: can we effectively utilize off-the-shelf general-purpose FMs and adapt them for medical report generation?

In this work, inspired by the state-of-the-art VL pre-training approach BLIP-2~\cite{li2023blip}, we present our encoder-decoder-based MRG model, where a lightweight query Transformer (Q-Former) was used to bridge two FMs: EVA ViT-g~\cite{fang2023eva} and ChatGLM-6B~\cite{zeng2023glm}, a bilingual LLM trained to align to human intentions. Different from BLIP-2, where the Q-Former was pre-trained with image-text pairs to bridge the modality gap of two FMs, we directly fine-tune the model on the target dataset. Furthermore, we perform ablative experiments on the trainable components of our MRG model to identify the crucial factors for effective transfer learning. Through experimentation on ImageCLEF 2023 caption prediction~\cite{ImageCLEFmedicalCaptionOverview2023}, we discover the significance of unfreezing EVA-ViT-g to learn medical image representations and parameter-efficient fine-tuning of ChatGLM-6B to capture the writing styles of medical reports. As a result, our best solution achieved the  4$^{th}$ and 2$^{nd}$ positions among 13 participating teams in the competition, as evaluated by BERTScore and ROUGE-1 metrics. 

\section{Related Works}

\paragraph{Medical Report Generation} Inspired by the success of neural-network-based image captioning~\cite{xu2015show,hossain2019comprehensive,hu2022scaling, yang2022clip,yang2023zeronlg}, medical report generation has attracted enormous research interest in recent years. Mainstream approaches for medical report generation adopt the \emph{de facto} encoder-decoder framework and focus on two aspects: 1) improving the cross-modal alignment between medical images and reports via reinforcement learning~\cite{li2018hybrid,qin2022reinforced}, architecture design~\cite{chen2020generating,you2021aligntransformer,chen2021cross}, or explicit loss constraints ~\cite{li2023unify,li2023dynamic} and 2) exploring retrieval- and knowledge-augmented report generation~\cite{li2018hybrid,liu2021exploring,li2023dynamic,li2019knowledge,zhang2020radiology}. These methods use dataset-related annotations to train their model, whose power, however, is restricted by the small number of labeled pairs.

\paragraph{Adaptation of Foundation Models} With more and more foundation models (FMs) springing up in CV, NLP, and their intersection, how to efficiently adapt FMs to a specific task becomes a research hotspot. One effective technique is \emph{prompt engineering}~\cite{liu2023pre}, which aims to affect the behaviors of language FMs by providing them with a textual template filled with task-related priors~\cite{raffel2020exploring,li2021prefix}, demonstrations of several examples~\cite{brown2020language,flamingo}, or a chain of thoughts~\cite{Wei2022CoT,kojima2022CoT,chowdhery2022palm}. Although such a technique usually does not require model optimization, it could be sensitive to the selection of templates and few-shot examples~\cite{lu2022fantastically,zhou2022learning}. Another popular technique is known as parameter-efficient transfer learning, which introduces lightweight components, e.g., Adapter~\cite{houlsby2019parameter}, continuous prompts~\cite{li2021prefix,liu2022p}, and LoRA~\cite{hulora}, into FMs to influence their activation and responses. With the aforementioned two techniques, recent practices have demonstrated the effectiveness of adapting general-purpose FMs to the medical domain~\cite{lievin2022can,singhal2022large,nori2023capabilities}. Different from these practices, we focus on medical report generation in this work.

\section{Approach}
\begin{figure}
  \centering
  \includegraphics[width=\linewidth]{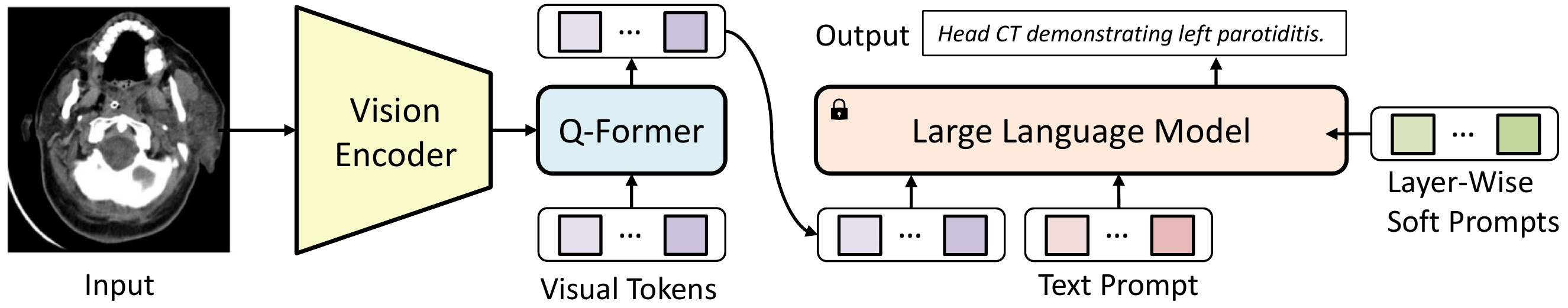}
  \caption{Overview of our model for medical report generation.}
  \label{fig:framework}
\end{figure}

\paragraph{Overview} As shown in Figure~\ref{fig:framework}, our model for medical report generation comprises three networks following BLIP-2~\cite{li2023blip}: a vision encoder, a query Transformer (Q-Former), and a LLM. To effectively transfer the knowledge of FMs, we unfreeze the vision encoder and parameter-efficiently fine-tune the language model with P-tuning~\cite{liu2022p}. 
Given the medical image $I$, the text prompt $\mathbf{x} = \{x_1, x_2, \dots, x_M\}$, and the target report $\mathbf{y} = \{y_1, y_2, \dots, v_T\}$, our model minimize the following negative log likelihood:
\begin{equation}
    \label{eq:loss_ce}
    \mathcal{L}(\theta) = -\sum_{t=1}^{T}\log p(y_t|\mathbf{y_{<t}}, \mathbf{x}, I; \theta),
\end{equation}
where $\theta$ denotes all parameters of the model, and $\mathbf{y_{<t}}$ means previously generated tokens before the $t$-th time step. Next, we introduce each component of our model.

\paragraph{Vision Encoder} We adopt EVA-ViT-g~\cite{fang2023eva} as the vision encoder to extract patch-based features $\boldsymbol{R} \in \mathbb{R}^{N\times d_v}$ of the medical image $I$. Since EVA-ViT-g uses a patch size of 14, we will obtain a long feature sequence for a high-resolution image, e.g., $N = 677$ for a $364\times364$ image.

\paragraph{Query Transformer} To reduce the computation costs of the subsequent modeling process, we adopt a Q-Former with $K$ ($K < N$) learnable visual tokens to obtain aggregated visual features. In the implementation, Q-Former is a BERT-like encoder \cite{devlin2018bert} with cross-attention layers inserted at a certain frequency.

\paragraph{Large Language Model} We adopt ChatGLM-6B~\cite{zeng2023glm} as the decoder to generate texts. Specifically, ChatGLM-6B is a decoder-only Transformer. Hence, we need to prefix the decoder input sequence with the aggregated visual features to achieve vision-grounded medical report generation. It is noted that ChatGLM-6B generates texts in Chinese or in English adaptively, i.e., no explicit signal is fed into the model to indicate which language to be generated.

\paragraph{P-Tuning} To preserve the ability of the LLM, we adopt the P-tuning~\cite{liu2022p} technique to fine-tune ChatGLM-6B. Specifically, the original self-attention layers of ChatGLM-6B follow the formulation below:
\begin{equation}
    Attn(\boldsymbol{Q}, \boldsymbol{K}, \boldsymbol{V}) = {\rm softmax}(\frac{\boldsymbol{Q}\boldsymbol{K}^T}{\sqrt{d_k}})\boldsymbol{V},
    \label{eq:attn}
\end{equation}
where $\boldsymbol{Q}$, $\boldsymbol{K}$, and $\boldsymbol{V}$ are queries, keys, and values mapped from the same input sequence. P-tuning introduces layer-wise soft prompts ($\boldsymbol{P_K}$ and $\boldsymbol{P_V}$) into the self-attention layers and modifies the keys and values. Thus, Equation.~\ref{eq:attn} is re-formulated as follows:
\begin{equation}
    Attn(\boldsymbol{Q}, {\rm Concat}(\boldsymbol{P_K}, \boldsymbol{K}), {\rm Concat}(\boldsymbol{P_V}, \boldsymbol{V})).
\end{equation}

\section{Experiments}
\begin{figure}
	\begin{minipage}[b]{0.32\columnwidth}
		\centering
		\subfloat[][]{\includegraphics[width=\linewidth]{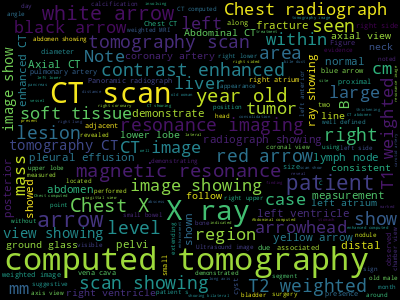}}
	\end{minipage}
    \begin{minipage}[b]{0.32\columnwidth}
		\centering
		\subfloat[][]{\includegraphics[width=\linewidth]{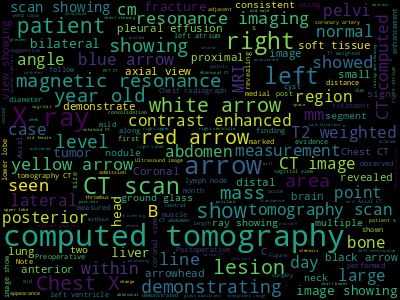}}
	\end{minipage}
	\begin{minipage}[b]{0.32\columnwidth}
		\centering
		\subfloat[][]{\includegraphics[width=\linewidth]{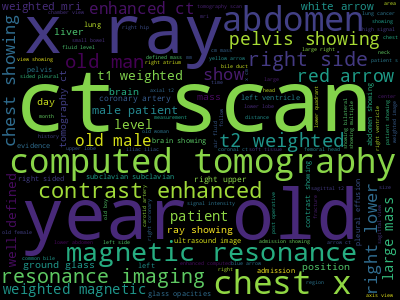}}
	\end{minipage}
	\caption{Word clouds of ground-truth reports from the training set (a) and the validation set (b). In (c), we show the word cloud of reports generated by our model on the validation set.}
 \label{fig:word_cloud}

\end{figure}

\subsection{Experimental Setups}

\paragraph{Dataset Statistics} ImageCLEF 2023 caption prediction is an updated and extended version of the Radiology Objects in COntext (ROCO) dataset~\cite{pelka2018radiology}. It provides 60,918, 10,437, and 10,473 radiology images for training, validation, and testing, respectively. Images are originated from biomedical articles and each image is annotated with one medical report. In Figure~\ref{fig:word_cloud} (a) and (b), we visualize the word clouds of ground-truth reports from the training and validation sets. As we can observe, there are many computed tomography (CT) and X-ray images in the dataset. Besides, there are some common expressions like ``white arrow'', indicating that a large portion of images have been marked by humans.

\paragraph{Metrics} We report two metrics following the guidelines of the competition\footnote{\url{https://www.imageclef.org/2023/medical/caption}}, namely BERTScore~\cite{zhang2019bertscore} and ROUGE-1~\cite{lin2004rouge}. Specifically, BERTScore is a model-based metric that calculates the semantic similarity of two sentences. ROUGE-1 measures the number of matching unigrams between a model-generated text and a reference.


\paragraph{Image Processing} During training, we process images with random resized cropping, horizontal flipping, and RandomAugment~\cite{cubuk2020randaugment}. During inference, we directly resize images to a specific resolution. We consider two image sizes in this paper: 224 and 364.

\paragraph{Model Settings} We use EVA-ViT-g~\cite{fang2023eva} as the vision encoder and v1.0 ChatGLM-6B\footnote{\url{https://github.com/THUDM/ChatGLM-6B}}~\cite{zeng2023glm} as the decoder by default. We always prompt ChatGLM-6B with the text ``Question: What is the radiology report for this image? Answer:" during training and inference. Following BLIP-2's official code\footnote{\url{https://github.com/salesforce/LAVIS}}, we implement BERT-like Q-Former, where cross-attention layers are inserted every two Transformer blocks and there are $K=32$ visual tokens to be learned. For P-tuning, we append 4 soft prompt tokens to the key and value sequences of each self-attention layer of ChatGLM-6B. Given that the number of layers and the model dimension of ChatGLM-6B is 28 and 4,096 respectively, P-tuning will introduce 0.9 M additional parameters. Besides ChatGLM-6B, we consider two more variants of language models: OPT-2.7B~\cite{zhang2022opt} and a randomly initialized Transformer-based decoder~\cite{vaswani2017attention} (abbr. TF-Base).

\paragraph{Hyper-Parameters} In the training phase, we truncate reports into a maximum length of 64. We use AdamW \cite{loshchilov2019decoupled} and L2 weight decay of 0.05 to train models with 12 samples per batch for 10 epochs. The learning rate is increased to 1e-4 after 2,000 warm-up steps and then follows a cosine annealing scheduler. After training, we use the beam search decoding algorithm with a beam size of 5 to generate medical reports. We set the repetition penalty to 2 to avoid duplication and limit the length of generated reports to at least 8.

\subsection{Results}
Table~\ref{tb:main} summarizes the performance of different model variants on ImageCLEF 2023 caption prediction. We have the following observations. 
\begin{itemize}
    \item Comparing \#\{1, 2, 3\}, we can see that using a frozen language FM is superior to training a small language model from scratch. This suggests that the syntactic knowledge of language FMs learned from common corpora can benefit medical report generation. 

    \item Comparing \#\{3, 4\}, we can observe that using the P-tuning technique enables ChatGLM-6B to generate more faithful medical reports, leading to 0.9\% absolute improvements under the ROUGE-1 metric.

    \item Comparing \#\{4, 5\}, we can see obvious improvements in both metrics after training the vision encoder together. This indicates the great differences between open-domain images and medical images and the necessity of medical image representation learning. 

    \item Comparing \#\{5, 6\}, we can conclude that a higher image resolution can preserve more details and thus benefits accurate medical report generation.
\end{itemize}
In short, customizing general-purpose FMs for medical report generation is feasible and can boost performance by a large margin (e.g., \#6 vs. \#1). Nevertheless, there still exist many potential improvement directions of our FM-based model, e.g., 
1) designing a multi-stage training procedure to learn from unlabeled medical images and texts and 2) speeding up the inference process via model compression and knowledge distillation. Additional scores (BLEU, BLEURT, METEOR, CIDEr, CLIPScore) for the competition can be found in the results folder~\cite{ImageCLEF2023} and on the website~\cite{ImageCLEFmedicalCaptionOverview2023}.

\begin{table*}[t]
\caption{Performance on ImageCLEF 2023 caption prediction. $\dagger$: Trainable components. $*$: P-tuning. Q-Former is omitted here, which is trainable and remains the same in all experiments.}
\label{tb:main}
\adjustbox{max width=\textwidth}{%
\setlength{\tabcolsep}{2pt}

\begin{tabular}{l l l l c cc cc cc}
\toprule

\multirow{2}{*}[-3pt]{\bf \#}
&\multirow{2}{*}[-3pt]{\begin{tabular}[l]{@{}l@{}}  \bf Image\\\bf Size\end{tabular}}
&\multirow{2}{*}[-3pt]{\begin{tabular}[l]{@{}l@{}}  \bf Vision\\\bf Encoder\end{tabular}}  
&\multirow{2}{*}[-3pt]{\begin{tabular}[l]{@{}l@{}}  \bf Language\\\bf Model\end{tabular}}
&\multicolumn{2}{c}{\bf \#Parameters}
&\multicolumn{2}{c}{\bf Validation Set}
&\multicolumn{2}{c}{\bf Test Set}
\\
\cmidrule(lr){5-6}
\cmidrule(lr){7-8}
\cmidrule(lr){9-10}
&&&
&Total &Trainable
&BERTScore &ROUGE-1
&BERTScore &ROUGE-1\\

\midrule

1   &224    &EVA-ViT-G  &TF-Base$\dagger$
&1.1B   &140.7M
&0.524705   &0.181943
&-  &-
\\

2   &224    &EVA-ViT-G  &OPT-2.7B
&2.8B   &105.2M
&0.599041   &0.230394
&-  &-
\\
3   &224    &EVA-ViT-G  &ChatGLM-6B
&7.3B   &108.3M
&0.606204   &0.231744
&-   &-
\\


4   &224    &EVA-ViT-G  &ChatGLM-6B$*$
&7.3B   &109.2M
&0.606173   &0.240584
&0.607918   &0.242249
\\



5   &224    &EVA-ViT-G$\dagger$  &ChatGLM-6B$*$
&7.3B   &1.1B
&0.617632   &0.263236
&-   &-\\

6 &364 &EVA-ViT-G$\dagger$ &ChatGLM-6B$*$
&7.3B   &1.1B
&\textbf{0.622668}   &\textbf{0.271068}
&\textbf{0.614836}   &\textbf{0.253279}\\


\bottomrule

\end{tabular}}

\end{table*}

\begin{figure}
  \centering
  \includegraphics[width=\linewidth]{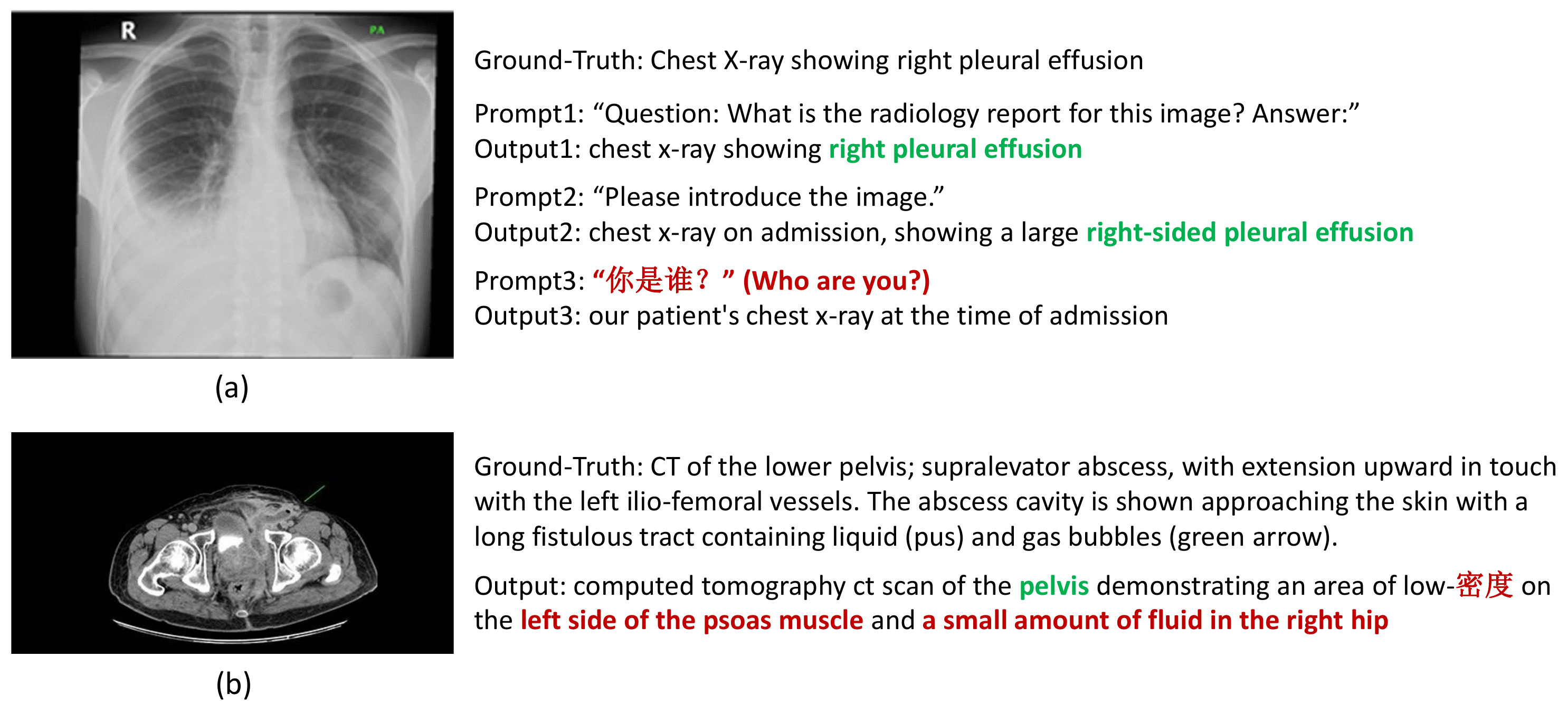}
  \caption{Two qualitative examples on the validation set of ImageCLEF 2023 caption prediction. Although our FM-based model may generate accurate keywords, it suffers from catastrophic forgetting of the question-answering ability, hallucination, and code-switching generation problems.}
  \label{fig:example}
\end{figure}

\subsection{Discussion}
We here give two qualitative examples in Figure~\ref{fig:example} to show three problems of our model.

\paragraph{Catastrophic Forgetting of the Question-Answering (QA) Ability} Our model is built upon ChatGLM-6B, a bilingual LLM with the QA ability. Considering that we keep ChatGLM-6B frozen all the time, it would be interesting to see how the QA ability is affected before and after training. As shown in Figure~\ref{fig:example} (a), when we change the prompt text from the default one (i.e., Prompt1) to the one irrelevant to the image content (i.e., Prompt3), the model's response does not meet with our expectation. In other words, we can not observe the ability of zero-shot visual QA as in the BLIP-2 paper~\cite{li2023blip}. We suspect such catastrophic forgetting is caused by two reasons: 1) model training does not include visual QA data, and 2) the prompt text fed into the model is fixed during training.

\paragraph{Hallucination} As shown in Figure~\ref{fig:example} (b), our model may produce  hallucinate unintended text. The word cloud in Figure~\ref{fig:word_cloud} (c) also shows that our model generates age information excessively. Such a hallucination problem is common among various natural language generation tasks~\cite{rohrbach2018object,yang2021non,ji2023survey}. One reason for this problem is that the cross-entropy loss for language modeling leads to the problem of exposure bias \cite{ranzato2015sequence}. To solve this problem, reinforcement learning is one of the effective techniques~\cite{rennie2017self,ouyang2022training}.

\paragraph{Code-Switching Generation} As shown in Figure~\ref{fig:example} (b), our model may generate Chinese tokens unexpectedly though it is fine-tuned on English-only data. On one hand, it indicates the risks to inherit the drawbacks of pre-trained models. On the other hand, it is a common phenomenon in multilingual machine translation systems~\cite{tang2020multilingual,costa2022no}. To overcome this problem, we may feed an explicit signal into the model to indicate which language to be generated or substitute the multilingual vocabulary with a monolingual one. Furthermore, we verify if the sporadic non-English tokens are semantically correct in Table~\ref{tab:chatgpt}, where we utilize ChatGPT to polish the generated reports with non-English tokens. As we can see, performance does not change basically. Therefore, there is still a huge room to improve our model.

\begin{table*}
  \caption{Effect of using ChatGPT to polish the generated reports with non-English tokens.}
  \label{tab:chatgpt}
\begin{tabular}{l cc}
\toprule

\multirow{2}{*}[-3pt]{\bf \# polished / total}
&\multicolumn{2}{c}{\bf Test Set}
\\
\cmidrule(lr){2-3}

&BERTScore &ROUGE-1\\

\midrule

0 / 10473 &0.614836   &\textbf{0.253279}\\

120 / 10473 &\textbf{0.615190}   &0.252756\\

\bottomrule
\end{tabular}
\end{table*}

\section{Conclusion}
In this work, we presented a simple yet effective attempt to customize general-purpose foundation models (FMs) for medical report generation. Our study not only shows the significant performance boost brought by FMs but also reveals the challenges of fine-tuning LLMs: catastrophic forgetting of the question-answering ability, hallucination, and code-switching generation. All these demerits should be addressed to develop a multilingual trustworthy chat-bot in a specific medical scenario. We leave this research to our future work. Source codes with model weights will be available at OpenMedIA\footnote{\url{https://openi.pcl.ac.cn/OpenMedIA}}~\cite{zhuang2022openmedia} after the conference.

\section*{Acknowledgments}
This work is partially supported by the National Natural Science Foundation of China (grant No. U21A20523 and 62176008 ). The computing resources of Pengcheng Cloudbrain are used in this research. We acknowledge the support provided by OpenI Community\footnote{\url{https://git.openi.org.cn}}. 

\bibliographystyle{unsrt}
\bibliography{main}
\end{document}